**Establishing Central Sensitization Inventory Cut-off Values in patients with Chronic Low Back Pain by Unsupervised Machine Learning**

Xiaoping Zheng[1], Claudine JC Lamoth[1], Hans Timmerman[2], Ebert Otten[1], and Michiel F Reneman[3*]

1. University of Groningen, University Medical Center Groningen, Department of Human Movement Sciences, Groningen, the Netherlands
2. University of Groningen, University Medical Center Groningen, Department of Anesthesiology, Pain Center, Groningen, the Netherlands
3. University of Groningen, University Medical Center Groningen, Department of Rehabilitation Medicine, Groningen, the Netherlands

* correspondence: m.f.reneman@umcg.nl

**Abstract:**

*Background:*

Central sensitization (CS) cannot be directly demonstrated in humans and it is proposed to refer to as Human Assumed Central Sensitization (HACS). HACS is involved in the development and maintenance of chronic low back pain (CLBP). Identifying HACS in individuals is crucial for tailoring appropriate treatment strategies, but there is no gold standard for assessing HACS. The Central Sensitization Inventory (CSI) was developed to evaluate the presence of HACS, with a cut-off value of 40/100 based on patients with chronic pain. However, various factors including pain conditions (e.g., CLBP, migraine, etc.), and gender may influence this cut-off value. For chronic pain condition such as CLBP, unsupervised clustering approaches can take these factors into consideration and automatically learn the HACS-related patterns. Therefore, this study aimed to determine the cut-off values for a Dutch-speaking population with CLBP, considering the total group and stratified by gender based on unsupervised machine learning.

*Methods:*

In this cross-sectional study, questionnaire data covering pain, physical, and psychological aspects were collected from patients with CLBP and aged-matched pain-free adults (referred to as healthy controls, HC). Four clustering approaches were applied to identify HACS-related clusters based on the questionnaire data and gender. The clustering performance was assessed using internal and external indicators. Subsequently, receiver operating characteristic (ROC) analysis was conducted on the best clustering results to determine the optimal cut-off values.

*Results*:

The study included 151 subjects, consisting of 63 HCs and 88 patients with CLBP. Hierarchical clustering yielded the best results, identifying three clusters: healthy group, CLBP with low HACS level, and CLBP with high HACS level groups. Based on the low HACS levels group (including HC and CLBP with low HACS level) and high HACS level group, the cut-off value for the overall groups were 35 (sensitivity 0.76, specificity 0.76), 34 for females (sensitivity 0.72, specificity 0.69), and 35 for males (sensitivity 0.92, specificity 0.81).

*Conclusion:*

The findings suggest that the optimal cut-off values for CLBP is 35. The gender-related cut-off values should be interpreted with caution due to the unbalanced gender distribution in the sample. The methodology employed in this study may provide new insights into identifying HACS-related patterns and contributes to establishing accurate cut-off values.

**Introduction:**

Central Sensitization (CS) refers to an increased responsiveness of nociceptive neurons in the central nervous system to their normal or subthreshold afferent input [1]. However, due to the inability to measure the mechanisms related to CS in individual humans [2], it has been proposed to refer to CS as Human Assumed Central Sensitization (HACS) [2]. HACS has been implicated in the development and maintenance of various chronic pain conditions, such as Chronic Low Back Pain (CLBP), fibromyalgia, and osteoarthritis [3]. CLBP is a leading contributor to global disability [4]. While the overall efficacy of rehabilitation for patients with CLBP is generally positive, the average effect sizes are modest [5]. The possible presence of HACS is one of the key factors contributing to the complexity of CLBP [6] which could be among the factors responsible for the modest treatment effects [7]. Recognizing HACS in individuals with CLBP is crucial for tailoring appropriate treatment strategies, as interventions targeting CS may differ from those addressing peripheral mechanisms [8, 9].

Despite its importance in recognizing HACS in CLBP, there is currently no universally accepted gold standard for diagnosing HACS [2]. The Central Sensitization Inventory (CSI) was developed as a self-report questionnaire to screen for the presence and severity of HACS in individuals experiencing musculoskeletal pain [10]. The CSI has demonstrated good psychometric properties in various pain conditions [11]. A cut-off value of 40 out of 100 was established based on a study involving patients with chronic pain with CS syndromes ((CSS), e.g., fibromyalgia, chronic fatigue syndrome, etc.), and it has demonstrated good sensitivity (81%) and specificity (75%) [12]. However, it has been observed that the cut-off values for the CSI vary across different types of musculoskeletal pain, ranging from 11 to 40 [12-15], as well as across different cultural and national contexts [16]. This variability highlights the need for establishing context-specific cut-off values to improve the utility of the CSI in diverse populations. Moreover, gender plays a significant role in pain, potentially affecting the presentation and severity of HACS and ultimately influencing the determination of the cut-off value [17, 18]. To the best of our knowledge, there is no cut-off value established for Dutch-speaking patients with CLBP. Because of the lack of gold standard, the relationship between the CSI and HACS remains ambiguous. It is uncertain whether the CSI indicates enhanced nociceptive responses or also a psychological hypervigilance [19, 20].

To address these challenges, unsupervised machine learning [21, 22] may be a possible approach. Unsupervised clustering approaches are data-driven, and can automatically learn the relationships between variables and explore the possible HACS-related subgroups based on the questionnaire data that reflect pain, physical functioning, psychological factors, and HACS. These clustering approaches do not rely on prior knowledge or assumptions about the underlying structure of the data and can identify distinct groups within the data based on patterns of HACS. Based on the clustering results, researchers can uncover the optimal cut-off value that best differentiates individuals with low and high levels of HACS in a data-driven and context-specific manner. Apart from this, these approaches are flexible and can be applied to various types of data [23], such as demographic, cultural, and psychosocial factors, making them suitable for analyzing the complex and multidimensional nature of HACS. Additionally, the good scalability [23] of these approaches makes them easily scalable to accommodate large datasets, such as electronic health record system. In the future, by collecting more diverse and representative samples of the Dutch-speaking patients with CLBP, this scalability ensures that the established cut-off value is robust and generalizable to the broader population with CLBP.

In this study, by using questionnaires which provide information about pain, physical, and psychological aspects, we aim to 1) explore the HACS-related subgroups based on unsupervised clustering approaches; 2) establish the optimal cut-off values of CSI within Dutch-speaking population with CLBP based on the clustering results; and examine gender differences in optimal cut-off values.

**Methods:**

*Participants*

The data of Dutch-speaking patients with CLBP utilized in the present study was extracted from an existing dataset of a broader study [24]. Data collection took place from September 2017 to September 2019, and comprehensive protocol details have been previously documented [24]. The aged-matched Dutch-speaking healthy controls (HC) were recruited by advertisements on social media and flyers.

The patients with CLBP were recruited from the outpatient Pain Rehabilitation Department at the Center for Rehabilitation of the University Medical Center Groningen (CvR-UMCG). CLBP is characterized by recurring pain in the lower back lasting for over 3 months. This pain is associated with emotional distress and/or functional disability and is not caused by any other diagnosis [25]. Inclusion criteria were as follows: 1) age ≥ 18 years; 2) admission to the interdisciplinary pain rehabilitation program; 3) ability to follow instructions; 4) signed informed consent. Patients were excluded if they: 1) had a specific diagnosis that better accounted for their CLBP symptoms (e.g., cancer, inflammatory diseases, or spinal fractures); 2) experienced neuralgia and/or radicular pain in the legs (examination by physiatrist); 3) were pregnant. The presence of comorbidities related to HACS (e.g., fibromyalgia, osteoarthritis or chronic fatigue syndrome) are no reason for exclusion from the study. The HCs were included if they: 1) were aged ≥ 18 years; 2) could follow instructions; 3) provided signed informed consent. Exclusion criteria for healthy controls: 1) report more than mild pain (evaluated by Visual Analogue Scale, see below); 2) use of antidepressant or antiepileptic drugs at the time of completing the questionnaire.

The Dutch-speaking patients with CLBP were collected with the approval of the Medical Research Ethics Committee of the University Medical Center Groningen (METc 2016/702). All procedures were conducted in accordance with the ethical principles outlined in the Declaration of Helsinki.

*Measures*

In this study, eight questionnaires assessed central HACS-related factors, including pain, physical functioning, psychological aspects, and HACS.

*Pain* was measured by Visual Analogue Scale (VAS), with values ranging from 0 to 100 mm. Values below 44 mm represent mild pain, 45 to 74 mm indicate moderate pain, and above 75 mm signify severe pain [26].

*Functioning* was evaluated using the Pain Disability Index (PDI) [27], the physical functioning subscale of the Rand36 questionnaire (Rand36-PF) [28], and the Work Ability Score (WAS) [29]. Higher PDI values (0-70) reflect greater pain interference with daily activities, while higher Rand36-PF values (0-100) indicate lower disability. WAS assessed self-reported workability, with higher values representing better workability.

*Psychological Aspects* were measured using the Pain Catastrophizing Scale (PCS, 0-52) [30], the Injustice Experience Questionnaire (IEQ, 0-48) [31], and the Brief Symptom Inventory (BSI global severity index t-score). PCS and IEQ values over 30 are clinically relevant, and higher BSI values denote more severe psychological symptoms.

*HACS* was evaluated by the CSI part A. CSI values can range from 0-100, with higher values assuming a higher level of CS [32]. Only section A was utilized in this study.

The data processing pipeline is depicted in Figure 1. Initially, questionnaire data and gender information were got from 151 subjects (Fig.1.a). Subsequently, the data were standardized using the Z-score approach, and Principal Component Analysis (PCA) was employed to reduce dimensionality (Fig.1.b). The first four components, accounting for 80% of the variance, were utilized. At the end, the features in each data sample which represents each subject were reduced from 9 to 4. Four kinds of clustering approaches were applied (Fig.1.c) to identify potential HACS-related groups. The optimal clusters were determined based on the most effective clustering results (Fig.1.d). Lastly, ROC analysis was conducted on these clusters to ascertain the best cut-off values for CSI (Fig.1.e).

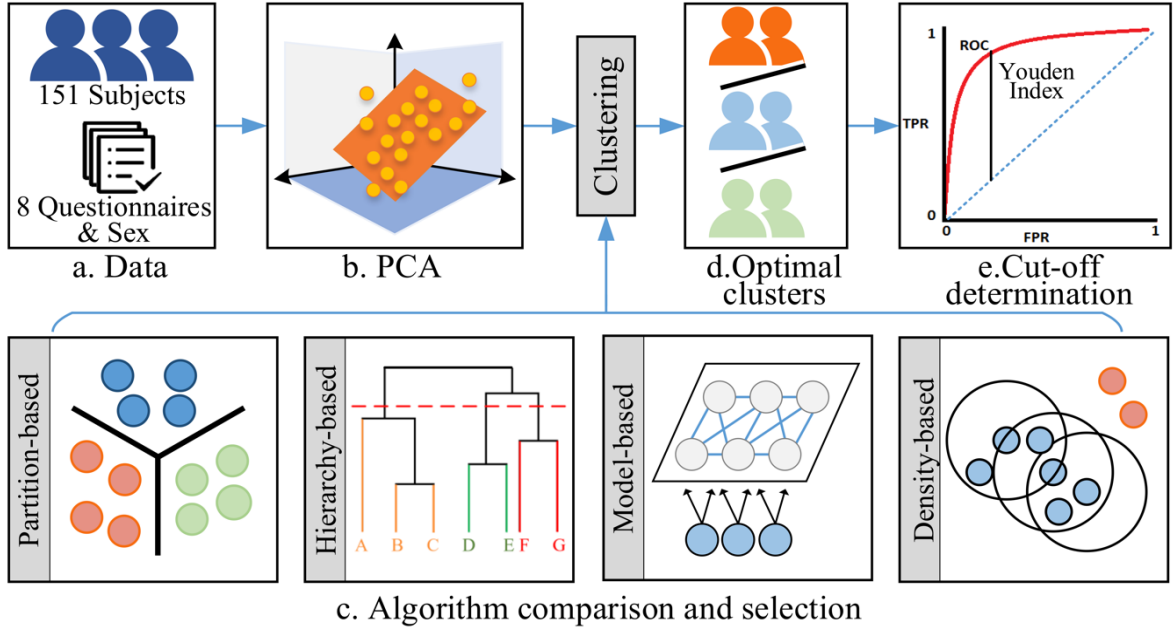

Figure 1. The data processing and analysis pipeline: (a) data collection; (b) PCA; (c) Clustering algorithms comparison and selection; (d) Optimal clusters; (e) Cut-off determination.

*Clustering approach*

After the pre-processing, to find the most suitable clustering approach for this study, 4 kinds of clustering approaches were included: K-means (partition based), Hierarchical clustering (hierarchy based), Self-organizing map (model-based), and Density-based spatial clustering of applications with noise (DBSCN) (Density-based), see Fig 1. c.

*K-Means* is one of the most commonly used clustering approaches based on partition [33]. The basic idea of this kind of clustering algorithm is to regard the center of data points as the center of the corresponding cluster. The algorithm can be summarized by the following procedure.

1) Randomly select $K$ data samples as the centroid of $K$ clusters and form a cluster prototype matrix as $M = [m_1, m_2, ..., m_k]$. In this study, the cluster number $K$ was set to 3.
2) Assign each data sample in the dataset to the nearest cluster.
3) Recalculate the centroid of the $K$ clusters based on the current partition and update the cluster prototype matrix $M' = [m'_1, m'_2, ..., m'_K]$..

$$m'_i = \frac{1}{|C_i|} \sum_{x \in C_i} x$$

where $C_i$ is the clusters $i$ with the centroid $m_1$, $x$ is the data sample in the dataset.

4) Repeat steps 2)-3) until there is no change for each cluster.

At the end, all data samples will be assigned to one of the three cluster to represent different HACS-related patterns. The K-means algorithm works well for compact and hyper spherical clusters.

*Hierarchical clustering* is a hierarchy-based clustering approaches, which constructs the hierarchical relationship among data in order to cluster [34]. In this study, clusters are formed by iteratively dividing the patterns using a bottom-up approach. The agglomerative clustering involved the following step can be summarized by the following procedure.

1) Start with $N$ atomic clusters which each of them includes exactly one data sample. $N$ represents the number of data samples of the whole data set, where $N = 151$. Calculate the Euclidean distance between any 2 clusters and form as a proximity matrix.

2) Search the minimal distance $d_{ij}$ in the proximity matrix and combine $C_i$ and $C_j$ to form a new cluster $C_{i \cup j}$.

3) Update the proximity matrix by computing the distances $d_{k(ij)}$ between the new cluster and the other clusters. $d_{k(ij)}$ can be computed by Lance–Williams algorithms as follow.

$$d_{k(ij)} = \alpha_i d_{ik} + \alpha_j d_{jk} + \beta d_{ij} + \gamma |d_{ik} - d_{jk}|$$

$$\alpha_i = \frac{n_i + n_k}{n_i + n_j + n_k}, \alpha_j = \frac{n_j + n_k}{n_i + n_j + n_k}, \beta = \frac{-n_k}{n_i + n_j + n_k}, \gamma = 0$$

where $n_i, n_j$ and $n_k$ are the sizes of the disjoint clusters $C_i$, $C_j$ and $C_k$.

4) Repeat steps 2)-3) until all objects are in the same clusters.

The results of hierarchical clustering are depicted by a dendrogram. The dendrogram describes the similarity between different data samples. The ultimate clustering results can be obtained by cutting the dendrogram at different levels.

*Self-organizing map* (SOM) is a model-based approach which is based on neural network learning approaches [35]. The core idea of SOM is to build a map of dimension reduction from the input space of high dimension to output space of low dimension on the assumption that there exists topology in the input data. The algorithm can be summarized by the following procedure.

1) Define the topology of the SOM with size $X * Y$.

$$X = Y = \sqrt{5\sqrt{N}}$$

where $N$ is the number of data samples (subjects) of the whole dataset and it is 151.

2) Randomly initialize the prototype weight matrix $W(0)_{X*Y*D}$ for the SOM network, where $D$ is the length of the features in the input data. In this study, $D$ is equal to 4 since the PCA reduce the dimension of each data sample from 9 to 4.

3) Calculate the distance between an input data sample $x$ and the nodes of network. The node $r_c$ which is closest to $x$ is chose as the winning node.

4) Update the weight matrix $W(t)$ as

$$w_{(i,j)}(t+1) = w_{(i,j)}(t) + \eta g_{(i,j)} \cdot (x - w_{(i,j)}(t))$$

where $w_{(i,j)}(t)$ means the weight of node at location $(i,j)$ at time $t$. $\eta g$ represents the neighborhood function and $\eta$ is the learning rate. $g_{(i,j)}$ is defined by the Gaussian method as

$$g_{(i,j)} = \exp\left(\frac{-||r_c - r_{(i,j)}||^2}{2\sigma^2}\right)$$

where $r_{(i,j)}$ is the node at the location $(i,j)$ in network.

5) Repeat steps 3)-4) until no change of neuron position that is more than a small positive number is observed.

*Density-based spatial clustering of applications with noise (DBSCAN)* is a clustering approach based on density, proposed by Ester Martin. It can find a cluster with any shape upon one density condition [36]. DBSCAN has the following basic concepts:

1) Set the radius of DBSCAN Algorithm Analysis neighbourhood as $\varepsilon$, and set the minimum number of data sample sets as $MinPts$. In this study, $\varepsilon$ and $MinPts$ were empirically determined and set to 15.

2) Randomly select one unvisited data sample $P$ and mark it as visited. If $N_\varepsilon(P) > MinPts$, then mark this data sample $P$ as core data sample. The $N_\varepsilon(P)$ is defined as

$$N_\varepsilon(P) = \{q \in D | dist(p,q) < \varepsilon\}$$

where $q$ is a data sample from the dataset $D$, $dist(p,q)$ means the distance between $P$ and $q$. If $dist(p,q) < \varepsilon$, $q$ and $P$ are directly density-reachable, such that $N_\varepsilon(P)$ is the number of data sample directly density-reachable from $P$.

3) Find out all the density-reachable data sample from $P$, mark them as visited and merge to the same cluster as $P$. The definition of density-reachable is: if there is a chain of objects $P_1, \ldots, P_n$, where $P_1 = P$, $P_n = Q$, and $P_{i+1}$ is directly density-reachable from $P_i$, such that $P$ is density-reachable from $Q$.

4) Repeat steps 2)-3), until all the objects are visited. The data samples which are visited but not in the clusters are noise data samples.

*Clustering performance evaluation*

To evaluate the clustering results of the unsupervised clustering approaches, the internal and external validation measures were used. Internal validation measures, including the silhouette coefficient, Davies-Bouldin index, and Calinski-Harabasz index, evaluate cluster quality based on clustering results. For external validation, clustering results were compared with known labels (HC). External validation assists in determining clustering accuracy and ensuring the meaningfulness of clustering outcomes.

*Silhouette Coefficient* indicates how cohesion of an object within its own cluster, and how separation of this object and other clusters [37]. The value closes to 1 means clusters are well apart from each other and clearly distinguished. The definition of the silhouette coefficient $SC$ is defined as:

$$SC = \max_K \tilde{s}(K)$$

where $K$ represents the number of clusters and $\tilde{s}(K)$ computes the mean value of $s(i)$ in cluster $K$. The definition of $s(i)$ is:

$$s(i) = \frac{b(i) - a(i)}{\max\{a(i), b(i)\}}$$

and,

$$a(i) = \frac{1}{|C_I| - 1} \sum_{j \in C_I, i \neq j} d(i,j)$$

$$b(i) = \min_{J \neq I} \frac{1}{|C_J|} \sum_{j \in C_J} d(i,j)$$

where $|C_I|$ represent the size of cluster $I$, $d(i,j)$ represents the distance between data sample $i$, and $j$.

*Davies-Bouldin index* is a function of the ratio of the within-cluster scatter, to the between-cluster separation [38]. A lower value will mean that the clustering is better. The scatter of cluster $I$ is defined as:

$$S_I = \frac{1}{|C_I|} \sum_{i=1}^{|C_I|} dis(i, A_I)$$

where $A_I$ is the centroid of $C_I$. The definition of separation of between clusters $I$ and $J$ is:

$$M_{I,J} = dis(A_I, A_j)$$

Davies-Bouldin index of $K$ clusters results can be computed as:

$$DB = \frac{1}{K} \sum_{I=1}^{K} \max_{I \neq J} \{\frac{S_I + S_J}{M_{i,J}}\}$$

The lower non-negative value of $DB$ means the better clustering results.

*Calinski-Harabasz index* is a measure of how similar a data sample is to its own cluster (cohesion) compared to other clusters (separation) [39]. The cohesion is estimated based on the distances from the objective in a cluster to its cluster centroid and the separation is based on the distance of the cluster centroids from the global centroid. This cohesion can be defined as:

$$CO = \frac{\sum_{I=1}^{K} |C_I| dis(A_I - A)}{K - 1}$$

where $A$ is the global centroid of the whole dataset.

The inter-cluster dispersion can be defined as:

$$SE = \frac{\sum_{I=1}^{K} \sum_{i}^{|C_I|} dis(i, A_I)}{N - K}$$

where $N$ is the number of data sample of the whole dataset.

Therefore, the Calinski-Harabasz index is:

$$CH = CO/SE$$

Higher value of CH index means the clusters are dense and well separated.

*Statistical Analyses*

In this study, the Mann–Whitney U test was applied to examine the differences in demographic characteristics. Based on the optimal clusters, the receiver operating characteristic (ROC) curve analysis [40] was used to suggest an optimal CSI cut-off value (shown in Fig 1. e.). The area under the ROC curve (AUC) represents the harmonic ratio of sensitivity and specificity. The Youden index was calculated to evaluate the performance of the accuracy of a diagnostic test. Positive predictive values (PPV) and negative predictive values (NPV) serve as valuable indicators of diagnostic accuracy, reflecting the proportion of true positives (corresponding to high level of HACS) and true negatives (corresponding with low level of HACS) among all positive and negative findings, respectively. These metrics contribute to a comprehensive understanding of the diagnostic performance. In addition to PPV and NPV, likelihood ratios are employed as critical statistical measures to assess the diagnostic efficacy of tests. The positive likelihood ratio (PLR) is calculated by dividing the true positive rate by the false positive rate. Similarly, the negative likelihood ratio (NLR) is determined by dividing the false negative rate by the true negative rate. These ratios provide insights into the ability of diagnostic tests to discriminate between individuals with different CS levels.

As a whole, the combined utilization of AUC, Youden-index, sensitivity, specificity, predictive values, and likelihood ratios was used to determine the optimal CSI cut-off values. For clinical use, the cut-off value should have a sensitivity plus specificity of at least 1.5, which is halfway between 1 (useless) and 2 (perfect) [41].

To ensure transparency and reproducibility of the findings, we have made the project repository publicly accessible at https://github.com/xzheng93/CSI_cutoff_establishment.

**Results:**

In this study, 296 subjects were included, while 139 subjects were excluded due to the incomplete questionnaires data, 6 subjects in HC group were excluded since they reported moderate pain. Therefore, 151 subjects (63 HC and 88 CLBP) were included in the data analysis. Table 1 shows the characteristics. The HC and CLBP groups are age-matched and were significantly different in BMI, but not in height and weight. The HC group reported less pain, better physical functioning, better psychological status, and lower CSI values. In terms of gender, females and males have matched BMI. Females are younger and smaller, and reported more pain, more disability, worse psychological status (e.g., depression, anxiety, distress, and pain catastrophising), and higher CSI values compared to male.

Table 1. Demography of participants

|  | HC (n=63) | CLBP (n = 88) | p-value | F(n=74) | M(n=77) | p-value |
|---|---|---|---|---|---|---|
| Gender | 23F/40M | 51F/37M | - | - | - | - |
| Age, years | 40.9 ± 13.5 | 41.4 ± 12.3 | =.988 | 37.5 ± 13.7 | 44.7 ± 10.7 | <.001 |
| Height, cm | 160.5 ± 56.7 | 175.2 ± 10.1 | =.085 | 158.6 ± 42.7 | 179.1 ± 29.9 | <.001 |
| Weight, kg | 83.6 ± 17.0 | 86.6 ± 16.9 | =.131 | 79.1 ± 16.3 | 91.3 ± 15.4 | <.001 |
| BMI, kg/m$^2$ | 25.8 ± 3.9 | 28.3 ± 5.6 | =.003 | 27.5 ± 6.2 | 27.0 ± 3.8 | =.64 |
| VAS (0–10) | 0.5 ± 0.7 | 4.2 ± 2.3 | <.001 | 3.3 ± 2.8 | 2.1 ± 2.3 | =.008 |
| PDI (0–70) | 5.1 ± 7.4 | 29.8 ± 14.4 | <.001 | 22.8 ± 16.1 | 16.3 ± 17.4 | =.009 |
| WAS (0–10) | 8.5 ± 1.3 | 4.9 ± 2.4 | <.001 | 5.9 ± 2.9 | 6.8 ± 2.3 | =.088 |
| Rand36-PF (0–100) | 28.4 ± 2.4 | 56.2 ± 21.4 | <.001 | 45.3 ± 20.4 | 44.0 ± 22.3 | =.39 |
| PCS (0–52) | 4.3 ± 4.8 | 15.5 ± 10.5 | <.001 | 13.6 ± 9.8 | 8.2 ± 9.9 | <.001 |
| IEQ (0–48) | 3.4 ± 5.5 | 14.1 ± 8.2 | <.001 | 12.5 ± 9.1 | 7.0 ± 7.8 | <.001 |
| BSI (t-score) | 32.5 ± 5.3 | 36.6 ± 9.4 | <.001 | 36.4 ± 8.4 | 33.5 ± 7.7 | =.004 |
| CSI (0–100) | 22.1 ± 9.7 | 38.1 ± 12.5 | <.001 | 34.5 ± 12.6 | 28.5 ± 14.4 | =.004 |

HC: Healthy Controls; CLBP: Chronic Low Back Pain; F: Female; M: Male; VAS: Visual Analogue Scale. BMI: Body mass index. PDI: Pain Disability Index. WAS: Work Ability Score. Rand36-PF: Rand 36-Physical Functioning subscale. PCS: Pain Catastrophizing Scale. IEQ: Injustice Experience Questionnaire. BSI: Brief Symptom Inventory, CSI: Central Sensitization Inventory.

After conducting 4 clustering approaches, their performance was compared and summarized in Table 2. With respect to internal indicators, hierarchical clustering, K-Means, and SOM demonstrated similar optimal values for Silhouette, Calinski-Harabasz, and Davies-Bouldin. In terms of external indicators, DBSACN clustered all the HC subjects (n=63) in the same cluster, but incorrectly classified 6 CLBP subjects within this cluster. Hierarchical clustering yielded a balanced outcome, clustering 62 HC subjects in the same cluster while misclassifying 3 CLBP subjects in the same cluster. Thus, hierarchical clustering may be considered the most suitable trade-off approach. Therefore, the results obtained from hierarchical clustering will be further analyzed to determine the optimal cut-off values of CSI.

Table 2. Clustering performance evaluation,

| Indicators | Approaches | Hierarchical Clustering | K-Means | DBSCAN | SOM |
| --- | --- | --- | --- | --- | --- |
| Internal | Silhouette | 0.47 | 0.48 | 0.34 | 0.47 |
|  | Calinski-Harabasz | 145.66 | 154.44 | 62.46 | 153.44 |
|  | Davies-Bouldin | 0.91 | 0.89 | 3.89 | 0.90 |
| External | True HC/Predicted HC | 62/65 | 60/65 | 63/69 | 60/63 |

DBSCAN: Density-based Spatial Clustering of Applications with Noise; SOM: Self-Organizing Map; HC: healthy controls.

Fig. 2 graphically demonstrates the clustering results of hierarchical clustering. On the y-axis, it represents the distances between distinct subjects and clusters. Meanwhile, the various colour blocks along the x-axis signify individuals from different groups, with red representing HC and blue representing CLBP. This figure distinctly demonstrates the clear separation between HC and CLBP. Within the CLBP cluster, two primary subgroups are distinguishable, represented by the colours grey and green. Consequently, this dendrogram implies the existence of three main clusters.

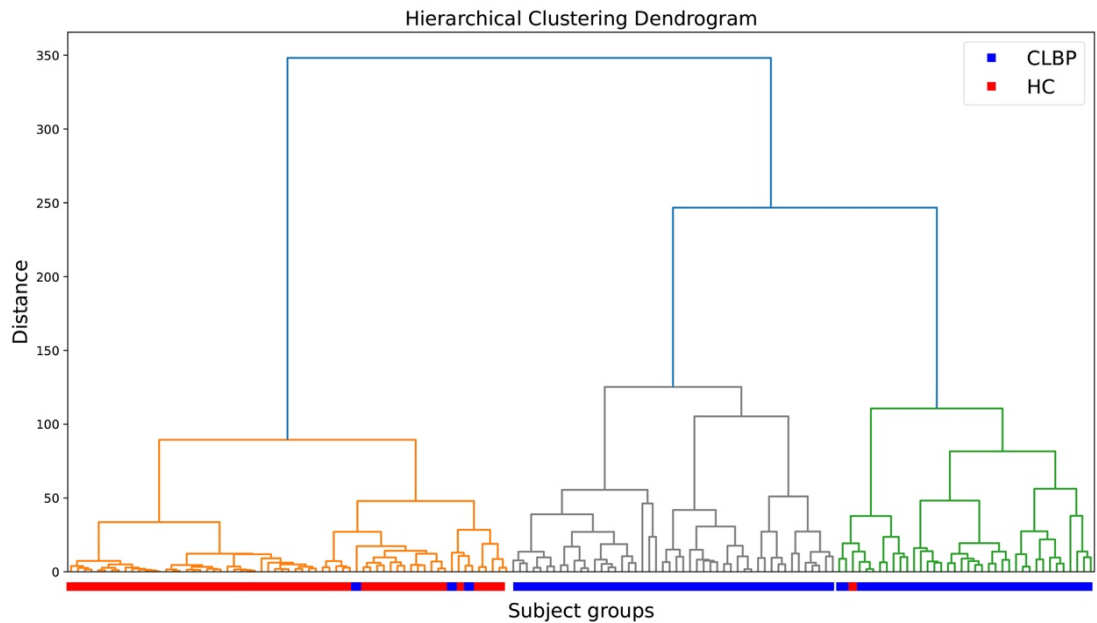

Figure 2. Dendrogram result of hierarchical clustering: red and blue blocks mean subjects from HC and CLBP groups, trees colored in orange, grey, and green indicate the presence of three primary clusters.

The CSI values for the three clusters identified by hierarchical clustering are depicted in Fig. 3 using a box plot. In this figure, red dots correspond to subjects from the HC group, while blue dots denote subjects from the CLBP group. Females are represented by round dots, while males are indicated by star dots. Additionally, green triangles are employed to indicate the mean CSI values of each box (cluster), and orange lines are used to indicate the median values. The boxes encapsulate the CSI values ranging from the first quartile to the third quartile, while the whiskers extend to show the minimum and maximum CSI values for each cluster. The box plot figures for other clustering approaches can be found in Appendix Figure A1, A2, and A3.

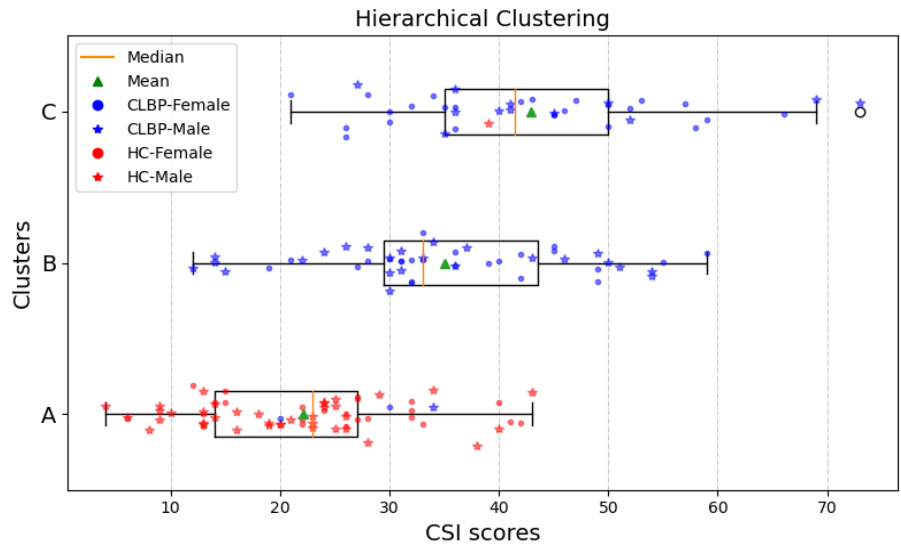

Figure 3. CSI clustering results of Hierarchical clustering. HC: Healthy Controls; CLBP: Chronic Low Back Pain; CSI: Central Sensitization Inventory.

Fig. 3 evidently shows that cluster A predominantly comprises most of the red dots (HC, N=62 out of 63) and a small number of blue dots (CLBP, N=3), indicating that cluster A may represent the healthy group. To understand the meaning of cluster B and C, the demographic characteristics of the hierarchical clustering results are displayed in Table 3. In comparison to cluster B, cluster C exhibited significantly higher levels of pain (VAS) and disability (PDI), lower work ability (WAS) and physical functioning (Rand36-PF), higher pain catastrophizing (PCS), injustice (IEQ), distress (BSI), and CSI values. Consequently, cluster C is characterized as patients with high HACS level, while cluster B represents patients with low HACS level. Hence, cluster A, B, and C represent the healthy, patients with low HACS levels, and patients with high CS levels groups.

Table 3. Demography of different clustering groups

|  | Cluster A | Cluster B | p-value of A & B | Cluster C | p-value of A & C | p-value of B & C |
|---|---|---|---|---|---|---|
| Gender | 25F/40M | 24F/24M | - | 25F/13M | - | - |
| Age, years | 41.2±13.6 | 39.4±11.8 | =.436 | 43.2±12.3 | =.77 | =.155 |
| Height, cm | 160.7±55.8 | 175.5±8.9 | =.232 | 175.4±11.6 | =.278 | =.807 |
| Weight, kg | 84.5±17.9 | 82.3±13.1 | =.9 | 90.5±18.6 | =.038 | =.028 |
| BMI, kg/m$^2$ | 26.3±4.8 | 26.8±4.1 | =.216 | 29.5±6.0 | =.002 | =.023 |
| VAS (0–10) | 0.5±0.9 | 3.3±1.9 | <.001 | 5.4±2.4 | <.001 | <.001 |
| PDI (0–70) | 5.3±7.2 | 22.9±13.2 | <.001 | 39.3±10.1 | <.001 | <.001 |
| WAS (0–10) | 8.4±1.4 | 6.0±2.0 | <.001 | 3.4±2.0 | <.001 | <.001 |
| Rand36-PF (0–100) | 28.7±2.6 | 71.1±14.4 | <.001 | 38.4±13.7 | <.001 | <.001 |
| PCS (0–52) | 4.5±4.8 | 11.0±8.1 | <.001 | 21.5±10.6 | <.001 | <.001 |
| IEQ (0–48) | 3.5±5.4 | 11.1±6.4 | <.001 | 18.3±8.5 | <.001 | <.001 |
| BSI (t-score) | 32.4±5.2 | 33.9±9.6 | =.038 | 40.5±8.0 | <.001 | =.001 |
| CSI (0–100) | 22.1±9.5 | 35.0±11.5 | <.001 | 42.9±12.2 | <.001 | =.007 |

F: Female; M: Male; VAS: Visual Analogue Scale. BMI: Body mass index. PDI: Pain Disability Index. WAS: Work Ability Score. Rand36-PF: Rand 36-Physical Functioning subscale. PCS: Pain Catastrophizing Scale. IEQ: Injustice Experience Questionnaire. BSI: Brief Symptom Inventory, CSI: Central Sensitization Inventory.

To establish the cut-off value for CSI to distinguish low and high levels of HACS, cluster A and B were combined to represent the low HACS levels population and cluster C was used to represent high HACS levels population. The demographic comparison of the overall low HACS and high HACS groups is presented in Table 4. Furthermore, to establish the cut-off values for females and males respectively, the females and males in the low HACS and high HACS groups were extracted for analysis, and the corresponding demographics are provided in Appendix Tables B1 and B2.

Table 4. Demography of low and high HACS samples

|  | Low HACS | High HACS | p-value |
|---|---|---|---|
| Gender | 49F/64M | 25F/13M | - |
| Age, years | 40.5±12.9 | 43.2±12.3 | =.357 |
| Height, cm | 167.0±43.3 | 175.4±11.6 | =.422 |
| Weight, kg | 83.6±16.1 | 90.5±18.6 | =.017 |
| BMI, kg/m$^2$ | 26.5±4.5 | 29.5±6.0 | =.002 |
| VAS (0–10) | 1.7±2.0 | 5.4±2.4 | <.001 |
| PDI (0–70) | 12.8±13.4 | 39.3±10.1 | <.001 |
| WAS (0–10) | 7.4±2.0 | 3.4±2.0 | <.001 |
| Rand36-PF (0–100) | 46.7±23.1 | 38.4±13.7 | =.301 |
| PCS (0–52) | 7.3±7.2 | 21.5±10.6 | <.001 |
| IEQ (0–48) | 6.8±6.9 | 18.3±8.5 | <.001 |
| BSI (t-score) | 33.0±7.4 | 40.5±8.0 | <.001 |
| CSI (0–100) | 27.6±12.2 | 42.9±12.2 | <.001 |

HACS: Human assumed central sensitization; F: Female; M: Male; VAS: Visual Analogue Scale. BMI: Body mass index. PDI: Pain Disability Index. WAS: Work Ability Score. Rand36-PF: Rand 36-Physical Functioning subscale. PCS: Pain Catastrophizing Scale. IEQ: Injustice Experience Questionnaire. BSI: Brief Symptom Inventory, CSI: Central Sensitization Inventory.

Based on the low and high levels HACS groups of overall, females, and males, ROC analysis was performed respectively, and the cut-off values for CSI, along with corresponding AUC, Youden Index, sensitivity, specificity, predictive values and likelihood ratio are presented in Table 5. In the table, the darker red colour represents better performance. By taken all the metrics into consideration, especially AUC and YI, the optimal cut-off value for the overall group is 35. Although cut-off values 34 and 35 for overall yielded the same AUC (=0.76) and YI (=0.52), but the cut-off value of 35 showed a more balance sensitivity and specificity. Therefore, the optimal cut-off value for the overall group is 35, with a AUC of 0.76, Youden Index of 0.52, sensitivity of 0.76, specificity of 0.76, PPV of 0.52, NPV of 0.91, PLR of 3.19, and NLR of 0.31. For females, the cut-off value remains 34 (AUC=0.71, Youden index=0.41, sensitivity=0.72, specificity=0.69, PPV=0.55, NPV=0.83, PLR=2.35, and NLR=0.4), while for males, the cut-off is 34 (AUC=0.87, Youden index=0.74, sensitivity=0.92, specificity=0.81, PPV=0.5, NPV=0.98, PLR=4.92, and NLR=0.09).
.

Table 5. CSI cut-off values

| | Overall | | | | | | | | Females | | | | | | | | Males | | | | | | | |
|---|---|---|---|---|---|---|---|---|---|---|---|---|---|---|---|---|---|---|---|---|---|---|---|---|
| CF | AUC | YI | Sen. | Spe. | PPV | NPV | PLR | NLR | AUC | YI | Sen. | Spe. | PPV | NPV | PLR | NLR | AUC | YI | Sen. | Spe. | PPV | NPV | PLR | NLR |
| 20 | 0.63 | 0.27 | 1 | 0.27 | 0.31 | 1 | 1.36 | 0 | 0.58 | 0.16 | 1 | 0.16 | 0.38 | 1 | 1.2 | 0 | 0.67 | 0.34 | 1 | 0.34 | 0.24 | 1 | 1.52 | 0 |
| 21 | 0.65 | 0.29 | 1 | 0.29 | 0.32 | 1 | 1.41 | 0 | 0.59 | 0.18 | 1 | 0.18 | 0.38 | 1 | 1.22 | 0 | 0.69 | 0.38 | 1 | 0.38 | 0.25 | 1 | 1.6 | 0 |
| 22 | 0.64 | 0.28 | 0.97 | 0.31 | 0.32 | 0.97 | 1.41 | 0.08 | 0.58 | 0.16 | 0.96 | 0.2 | 0.38 | 0.91 | 1.21 | 0.2 | 0.7 | 0.39 | 1 | 0.39 | 0.25 | 1 | 1.64 | 0 |
| 23 | 0.66 | 0.32 | 0.97 | 0.35 | 0.33 | 0.98 | 1.49 | 0.08 | 0.6 | 0.2 | 0.96 | 0.24 | 0.39 | 0.92 | 1.27 | 0.16 | 0.71 | 0.42 | 1 | 0.42 | 0.26 | 1 | 1.73 | 0 |
| 24 | 0.67 | 0.35 | 0.97 | 0.37 | 0.34 | 0.98 | 1.55 | 0.07 | 0.6 | 0.2 | 0.96 | 0.24 | 0.39 | 0.92 | 1.27 | 0.16 | 0.73 | 0.47 | 1 | 0.47 | 0.28 | 1 | 1.88 | 0 |
| 25 | 0.69 | 0.39 | 0.97 | 0.42 | 0.36 | 0.98 | 1.67 | 0.06 | 0.61 | 0.23 | 0.96 | 0.27 | 0.4 | 0.93 | 1.31 | 0.15 | 0.77 | 0.53 | 1 | 0.53 | 0.3 | 1 | 2.13 | 0 |
| 26 | 0.71 | 0.42 | 0.97 | 0.44 | 0.37 | 0.98 | 1.75 | 0.06 | 0.61 | 0.23 | 0.96 | 0.27 | 0.4 | 0.93 | 1.31 | 0.15 | 0.79 | 0.58 | 1 | 0.58 | 0.32 | 1 | 2.37 | 0 |
| 27 | 0.7 | 0.41 | 0.92 | 0.49 | 0.38 | 0.95 | 1.79 | 0.16 | 0.59 | 0.19 | 0.88 | 0.31 | 0.39 | 0.83 | 1.27 | 0.39 | 0.81 | 0.62 | 1 | 0.62 | 0.35 | 1 | 2.67 | 0 |
| 28 | 0.71 | 0.42 | 0.89 | 0.52 | 0.39 | 0.94 | 1.87 | 0.2 | 0.63 | 0.27 | 0.88 | 0.39 | 0.42 | 0.86 | 1.44 | 0.31 | 0.77 | 0.55 | 0.92 | 0.62 | 0.33 | 0.98 | 2.46 | 0.12 |
| 29 | 0.71 | 0.43 | 0.87 | 0.56 | 0.4 | 0.93 | 1.96 | 0.24 | 0.63 | 0.27 | 0.84 | 0.43 | 0.43 | 0.84 | 1.47 | 0.37 | 0.79 | 0.58 | 0.92 | 0.66 | 0.35 | 0.98 | 2.69 | 0.12 |
| 30 | 0.72 | 0.43 | 0.87 | 0.57 | 0.4 | 0.93 | 2 | 0.23 | 0.63 | 0.27 | 0.84 | 0.43 | 0.43 | 0.84 | 1.47 | 0.37 | 0.8 | 0.59 | 0.92 | 0.67 | 0.36 | 0.98 | 2.81 | 0.11 |
| 31 | 0.71 | 0.43 | 0.82 | 0.61 | 0.41 | 0.91 | 2.1 | 0.3 | 0.61 | 0.23 | 0.76 | 0.47 | 0.42 | 0.79 | 1.43 | 0.51 | 0.82 | 0.64 | 0.92 | 0.72 | 0.4 | 0.98 | 3.28 | 0.11 |
| 32 | 0.73 | 0.46 | 0.82 | 0.65 | 0.44 | 0.91 | 2.3 | 0.29 | 0.64 | 0.27 | 0.76 | 0.51 | 0.44 | 0.81 | 1.55 | 0.47 | 0.84 | 0.67 | 0.92 | 0.75 | 0.43 | 0.98 | 3.69 | 0.1 |
| 33 | 0.74 | 0.49 | 0.79 | 0.7 | 0.47 | 0.91 | 2.62 | 0.3 | 0.68 | 0.35 | 0.72 | 0.63 | 0.5 | 0.82 | 1.96 | 0.44 | 0.84 | 0.67 | 0.92 | 0.75 | 0.43 | 0.98 | 3.69 | 0.1 |
| 34 | 0.76 | 0.52 | 0.79 | 0.73 | 0.5 | 0.91 | 2.97 | 0.29 | <u>0.71</u> | <u>0.41</u> | <u>0.72</u> | <u>0.69</u> | <u>0.55</u> | <u>0.83</u> | <u>2.35</u> | <u>0.4</u> | 0.84 | 0.69 | 0.92 | 0.77 | 0.44 | 0.98 | 3.94 | 0.1 |
| 35 | <u>0.76</u> | <u>0.52</u> | <u>0.76</u> | <u>0.76</u> | <u>0.52</u> | <u>0.91</u> | <u>3.19</u> | <u>0.31</u> | 0.69 | 0.37 | 0.68 | 0.69 | 0.53 | 0.81 | 2.22 | 0.46 | <u>0.87</u> | <u>0.74</u> | <u>0.92</u> | <u>0.81</u> | <u>0.5</u> | <u>0.98</u> | <u>4.92</u> | <u>0.09</u> |
| 36 | 0.74 | 0.47 | 0.71 | 0.76 | 0.5 | 0.89 | 2.97 | 0.38 | 0.67 | 0.33 | 0.64 | 0.69 | 0.52 | 0.79 | 2.09 | 0.52 | 0.83 | 0.66 | 0.85 | 0.81 | 0.48 | 0.96 | 4.51 | 0.19 |
| 37 | 0.7 | 0.39 | 0.61 | 0.79 | 0.49 | 0.86 | 2.85 | 0.5 | 0.65 | 0.29 | 0.56 | 0.73 | 0.52 | 0.77 | 2.11 | 0.6 | 0.76 | 0.52 | 0.69 | 0.83 | 0.45 | 0.93 | 4.03 | 0.37 |
| 38 | 0.7 | 0.4 | 0.61 | 0.8 | 0.5 | 0.86 | 2.97 | 0.5 | 0.65 | 0.29 | 0.56 | 0.73 | 0.52 | 0.77 | 2.11 | 0.6 | 0.77 | 0.54 | 0.69 | 0.84 | 0.47 | 0.93 | 4.43 | 0.36 |
| 39 | 0.71 | 0.41 | 0.61 | 0.81 | 0.51 | 0.86 | 3.11 | 0.49 | 0.65 | 0.29 | 0.56 | 0.73 | 0.52 | 0.77 | 2.11 | 0.6 | 0.78 | 0.55 | 0.69 | 0.86 | 0.5 | 0.93 | 4.92 | 0.36 |
| 40 | 0.7 | 0.39 | 0.58 | 0.81 | 0.51 | 0.85 | 3.12 | 0.52 | 0.66 | 0.32 | 0.56 | 0.76 | 0.54 | 0.77 | 2.29 | 0.58 | 0.74 | 0.47 | 0.62 | 0.86 | 0.47 | 0.92 | 4.38 | 0.45 |
| 41 | 0.7 | 0.39 | 0.55 | 0.84 | 0.54 | 0.85 | 3.47 | 0.53 | 0.68 | 0.36 | 0.56 | 0.8 | 0.58 | 0.78 | 2.74 | 0.55 | 0.71 | 0.41 | 0.54 | 0.88 | 0.47 | 0.9 | 4.31 | 0.53 |
| 42 | 0.67 | 0.35 | 0.5 | 0.85 | 0.53 | 0.83 | 3.32 | 0.59 | 0.69 | 0.38 | 0.56 | 0.82 | 0.61 | 0.78 | 3.05 | 0.54 | 0.63 | 0.26 | 0.38 | 0.88 | 0.38 | 0.88 | 3.08 | 0.7 |
| 43 | 0.67 | 0.35 | 0.47 | 0.88 | 0.56 | 0.83 | 3.82 | 0.6 | 0.7 | 0.4 | 0.52 | 0.88 | 0.68 | 0.78 | 4.25 | 0.55 | 0.63 | 0.26 | 0.38 | 0.88 | 0.38 | 0.88 | 3.08 | 0.7 |
| 44 | 0.67 | 0.34 | 0.45 | 0.89 | 0.59 | 0.83 | 4.21 | 0.62 | 0.68 | 0.36 | 0.48 | 0.88 | 0.67 | 0.77 | 3.92 | 0.59 | 0.65 | 0.29 | 0.38 | 0.91 | 0.45 | 0.88 | 4.1 | 0.68 |
| 45 | 0.67 | 0.34 | 0.45 | 0.89 | 0.59 | 0.83 | 4.21 | 0.62 | 0.68 | 0.36 | 0.48 | 0.88 | 0.67 | 0.77 | 3.92 | 0.59 | 0.65 | 0.29 | 0.38 | 0.91 | 0.45 | 0.88 | 4.1 | 0.68 |

CF: cut-off values; AUC: area under the curve; YI: Youden index; Spe: specificity; Sen: sensitivity; PPV: positive predictive values; NPV: negative predictive values; PLR: positive likelihood ratio; NLR: negative likelihood ratio. The optimal cut-off values are underlined. The darker red colour represents better performance.

**Discussion:**

The aim of the present study was to explore HACS-related subgroups via unsupervised clustering approaches based on questionnaires data and establish CSI cut-off values for Dutch-speaking population with CLBP. The clustering results showed three distinct clusters: healthy group, patients with low HACS levels, and patients with high HACS levels. These clusters exhibited variations in pain intensity, disability levels, and psychological status. By comparing the low HACS level individuals (including healthy group), ROC analysis indicated the optimal cut-off values of 35 in the total group and for males, and 34 for females.

To evaluate the performance of the clustering algorithms, both external and internal metrics were utilized in the present study. The clustering outcomes demonstrated that, across all methods, the majority of HC subjects were grouped into the same cluster. It may suggest that the proposed clustering approaches were accurate to a certain degree. Internal metrics evaluate the separation and cohesion of clusters; and, the values of these indicators may not support the result that clusters were well-separated. This might be attributed to the inherent nature of HACS, which is not strictly binary in its essence. Rather, HACS likely exists along a continuum, spanning from absent to more pronounced degrees [42]. The demographics of the clustering results reveal that cluster C (high HACS level group) exhibited the most severe pain, greatest disability, and poorest psychological status (e.g., depression, anxiety, distress, and pain catastrophising). In contrast, cluster B (patients with CLBP and low HACS level) occupies the middle of the spectrum, while cluster A (healthy group) is situated at the opposite end.

The cutoff value for the CSI in our study for Dutch-speaking patients with CLBP is established at 35, with an AUC of 0.76, Youden Index of 0.52, sensitivity of 0.76, and specificity of 0.76. Three other studies have established CSI cutoff values for Dutch-speaking patients with chronic pain [18, 43, 44]. Initially, the CSI was translated into Dutch, and the Dutch version demonstrated sufficient test-retest reliability (ICC=0.88) internal consistency (Cronbach's alpha=0.91), and appropriate structural validity [44]. This study recommended employing a cutoff value of 40 for identifying patients (not solely CLBP) at risk of exhibiting signs of HACS based on earlier research [12]. The determination of this cutoff value of 40 was based solely on patients with chronic pain and CSS, as well as HCs, while patients without CSS were excluded [12]. It yielded a sensitivity of 0.81 and specificity of 0.75. A recent study established the cut-off values 30 for Dutch-speaking patients with chronic pain and at least one CSS, compared to HCs, reporting high sensitivity (0.85) and specificity (0.92) [18]. However, because patients with chronic pain and without CSS were excluded, the sensitivity and specificity scores of the cut-off values may not accurately reflect the discriminative ability between patients with or without CSS. A follow-up study [45] employed the cutoff value of 40 to distinguish between patients with chronic pain with or without CSS, showing similar sensitivity (0.83), but a notably decreased specificity of 0.55. As demonstrated in our study, there are distinct differences in pain intensity, disability levels, and psychological status between HCs and patients with chronic pain and high levels of HACS, whereas patients with chronic pain and low levels of HACS fall in the middle. In our study we conducted a sensitivity analysis by comparing cluster C (comprising patients with high HACS levels) exclusively with the HC group, as elaborated in Appendix C. Remarkably, the optimal cutoff values remained consistent, and, simultaneously, all metrics exhibited substantial improvements (AUC=0.83, Youden Index=0.65, sensitivity=0.76, and specificity=0.89). These results may suggest the need to include patients with low levels of HACS when determining and evaluating the cutoff values. Another research, which established cutoff values for Dutch-speaking patients with chronic pain, identified four clinically relevant categories: low (0–26), mild (27–39), moderate (40–52), and high (53+) [43]. However, this study only utilized the CSI value distribution of patients with chronic pain while excluding HCs. The distribution may change and vary based on the assessed population. Apart from this, this approach may lead to suboptimal cutoff values since it does not allow for discrimination between patients with HACS, as well as HCs. Therefore, in our study, the optimal cut-off values of CSI were determined based on high HACS levels and low HACS levels (with HC) groups.

Because no gold standard exists for assessing HACS, previous studies have employed various methods to indirectly determine the presence of HACS. Some studies assume that the presence of HACS can be indicated by one or several CSS [12, 15, 18]. The presence of CSS was assessed by a physician based on symptom complaints or thorough physical examination (e.g., tender-point evaluations fibromyalgia) [46] or self-reported questions (such as section B of the CSI which asks participants if they have been diagnosed with CSS) [12, 18]. However, due to the lack of a gold standard, expert judgment may vary across clinicians, potentially introducing bias into the cut-off value [12]. Apart from this, HACS can exist even when a CSS is absent [6]. Some studies use quantitative sensory testing (QST) to evaluate the dynamic modulation of nociceptive signals [47] as an indicator for the presence of HACS based on pressure pain threshold, temporal summation, conditioned pain modulation, and thermal QST [14, 48, 49]. However, QST is time-consuming and requires specialized equipment and trained personnel, and does not take the physical functioning and psychosocial issues into consideration while HACS is also related to these factors [50, 51]. Moreover, there is an absence of established cut-off values for QST for the assessment of HACS [2]. Our study employed data-driven clustering approaches to automatically uncover potential patterns in individuals based on pain, physical functioning, and psychological factors. Through the examination of the interrelationships among these factors, the clustering results indicated the division of the CLBP group and HC group into three primary clusters (cluster A, B and C). Their demographic results may support the notion that cluster B and C are associated with HACS, as they exhibit significant differences in pain, physical functioning, and psychological status compared with healthy group (cluster A). In addition to this, our study, through the association of CSI values and psychological states across the three clusters, may corroborate the finding that CSI is associated with psychological constructs [19]. However, since no biological measures were included in our study, it is not possible to determine if CSI is exclusively associated with psychological constructs.

Literature indicates significant differences in pain perception between genders [52-54]. Females generally exhibit a higher prevalence of clinical pain disorders and lower pain thresholds compared to males [52-54]. Accordingly, it was expected that females would demonstrate higher cut-off values for the CSI, indicating higher sensitivity to nociceptive stimuli. Several studies have provided evidence in support of this hypothesis [45, 55, 56]. In our study, females reported higher levels of pain, disability, CSI values, and worse psychological status. However, this could be confounded by the higher number of females in the CLBP group. Contrary to previous research [55, 56], our study did not find higher cut-off value for females compared to males (34 vs. 35) , and the sum of our sensitivity and specificity for the females group was 1.41, below 1.5 [41]. Given the unequal distribution of male and female participants in both CLBP and HC groups, the distributions of their CSI values are also uneven. Hence, the gender-related cut-off values derived from this study should be interpreted cautiously.

**Strengthen and Limitation:**

The utilization of clustering approaches offers a flexible and adaptable methodology that can accommodate diverse data types, thereby providing valuable insights into the complex and multidimensional characteristics of HACS. As knowledge of HACS continues to expand, the methodology employed in this study can be applied to identify patterns associated with HACS, incorporating increasingly precise factors that accurately capture the essence of HACS. Furthermore, these clustering approaches can be implemented within the electronic health record system. As the system expands, the scalability of the clustering approaches allows for learning from each case, leading to the generation of increasingly robust cut-off values. This contributes to the advancement of 'Data Driven Health Care'.

There were several limitations to the current study. Firstly, the data for our study were obtained from a larger study with different objectives. As a result, some critical information, such as part B of the CSI and objective measurements like QST, was absent. These missing details could provide insights into changes in pain thresholds and the presence of widespread pain, which could suggest alterations in central pain processing mechanisms [57] and possibly indicative of HACS. However, once we acquired these additional pieces of information, the flexibility and adaptability of the proposed approach enabled us to redo the analysis easily, thereby identifying patterns related to HACS and determining more accurate CSI cutoff values. Secondly, in our study, levels of HACS were clustered using unsupervised machine learning based on questionnaire results and gender. While cluster C exhibited patterns associated with high levels of HACS, characterized by increased pain, poorer physical and psychological conditions, and higher CSI values, it is important to note that individuals in cluster C should not be directly classified as patients with HACS, and individuals in cluster B should not be assumed to be without HACS. Instead, the clustering results provide insights into the severity of HACS levels among participants [32]. It is worth acknowledging that utilizing CSI as an input for identifying optimal clustering results and determining the best cut-off values for CSI introduces the potential risk of circular reasoning. However, the results of the feature importance analysis using PCA (further details provided in Appendix D. Fig. 1) indicate that the clustering process primarily relied on information such as gender, Rand-36, and BSI, with CSI ranking 7 out of 9 variables. Consequently, the associated risk is minimal. Thirdly, our study did not exclude other pain conditions due to the complex nature of chronic pain and the lack of comprehensive documentation. It is plausible that HACS may not be solely attributed to CLBP, and consequently, the CSI cut-off values established by our study may apply to patients with CLBP in combination with other pain conditions. Lastly, the cut-off value for females may not be valid enough since the sensitivity pluses specificity lower than 1.5. Apart from this, the distribution in gender was imbalanced between the CLBP and HC groups. Consequently, the proposed cut-off values for the genders separately should be interpreted with caution.

**Conclusion**

This study explored the HACS-related clusters based on pain intensity, disability levels, psychological status, and gender. Three distinct cluster were found by the data-driven approaches. patients with high HACS levels group and healthy group were at 2 ends, while patients with low HACS levels group were at the middle. A cut-off value of 35 on the CSI for the Dutch-speaking population with CLBP was established, aiming to differentiate between low and high levels of HACS.

The methodology employed in this study offers a data-driven means to identify subgroups, establish optimal diagnostic thresholds, and enriches the comprehension of this intricate HACS. Ultimately, this methodology empowers researchers and clinicians to craft more personalized and effective approaches for assessing and managing conditions associated with HACS and other diseases.

31. Bults, R.M., et al., *Test-retest reliability and construct validity of the dutch injustice experience questionnaire in patients with chronic pain.* Psychological Injury and Law, 2020. **13**: p. 316-325.

32. Neblett, R., et al., *Establishing clinically relevant severity levels for the central sensitization inventory.* Pain Practice, 2017. **17**(2): p. 166-175.

33. Hamerly, G. and C. Elkan, *Learning the k in k-means.* Advances in neural information processing systems, 2003. **16**.

34. Murtagh, F. and P. Contreras, *Algorithms for hierarchical clustering: an overview.* Wiley Interdisciplinary Reviews: Data Mining and Knowledge Discovery, 2012. **2**(1): p. 86-97.

35. Kohonen, T., *The self-organizing map.* Proceedings of the IEEE, 1990. **78**(9): p. 1464-1480.

36. Schubert, E., et al., *DBSCAN revisited, revisited: why and how you should (still) use DBSCAN.* ACM Transactions on Database Systems (TODS), 2017. **42**(3): p. 1-21.

37. Aranganayagi, S. and K. Thangavel. *Clustering categorical data using silhouette coefficient as a relocating measure.* in *International conference on computational intelligence and multimedia applications (ICCIMA 2007).* 2007. IEEE.

38. Petrovic, S. *A comparison between the silhouette index and the davies-bouldin index in labelling ids clusters.* in *Proceedings of the 11th Nordic workshop of secure IT systems.* 2006. Citeseer.

39. Wang, X. and Y. Xu. *An improved index for clustering validation based on Silhouette index and Calinski-Harabasz index.* in *IOP Conference Series: Materials Science and Engineering.* 2019. IOP Publishing.